\documentclass[letterpaper, 10 pt, conference]{ieeeconf} 
\IEEEoverridecommandlockouts                              % This command is 
\usepackage{ifpdf}
\usepackage{cite}
\usepackage[hyphens]{url}
\usepackage{graphicx}
\usepackage{booktabs}
\usepackage{threeparttable}
\usepackage[cmex10]{amsmath}
\usepackage{mathtools}% 
\usepackage{algorithmic}
\usepackage{array}
\usepackage{stfloats}
\usepackage[switch]{lineno}
\usepackage{amsfonts}
\usepackage{color}
\usepackage{pifont}
\usepackage{hhline}
\usepackage{colortbl}
\definecolor{Gray}{gray}{0.8}
\usepackage[]{footmisc}
\usepackage{soul}

\soulregister\cite7
\soulregister\ref7
\soulregister\pageref7
\title{\LARGE \bf Trimanipulation: Evaluation of human performance \\ in a 3-handed coordination task}

\author{Yanpei~Huang$^{1,2*}$, Jonathan~Eden$^{1*}$, Ekaterina~Ivanova$^1$,  Soo~Jay~Phee$^2$ and Etienne~Burdet$^{1*}$ 
 
\thanks{This work was funded by the Singapore National Research Foundation through the NRF Investigatorship Award (NRF-NRFI 2016-07), by the UK EPSRC FAIRSPACE, and EU H2020 PH-CODING (FETOPEN 829186), TRIMANUAL (MSCA 843408), NIMA (FETOPEN 899626) grants.}
\thanks{$^{1}$Authors are with the Department of Bioengineering, Imperial College of Science Technology and Medicine, London, UK. $^2$Authors are or were with the School of Mechanical and Aerospace Engineering, Nanyang Technological University, Singapore. $^*$Corresponding authors: \{yanpei.huang, j.eden, e.burdet\}@imperial.ac.uk.}}

\begin{document}
\maketitle
\thispagestyle{empty}
\pagestyle{empty}

\begin{abstract}
Many teleoperation tasks require three or more tools working together, which need the cooperation of multiple operators. The effectiveness of such schemes may be limited by communication. Trimanipulation by a single operator using an artificial third arm controlled together with their natural arms is a promising solution to this issue. Foot-controlled interfaces have previously shown the capability to be used for the continuous control of robot arms. However, the use of such interfaces for controlling a supernumerary robotic limb (SRLs) in coordination with the natural limbs, is not well understood. In this paper, a teleoperation task imitating physically coupled hands in a virtual reality scene was conducted with 14 subjects to evaluate human performance during trimanipulation. The participants were required to move three limbs together in a coordinated way mimicking three arms holding a shared physical object. It was found that after a short practice session, the three-hand trimanipulation using a single subject's hands and foot was still slower than dyad operation, however, they displayed similar performance in success rate and higher motion efficiency than two person's cooperation. 

\end{abstract}

\section{Introduction}
Teleoperation is a key solution for interacting with remote objects and in dangerous environments \cite{9134968,8890885}. Many teleoperation tasks require more than two instruments for successful task completion, e.g in robotic laparoscopic surgery, three instruments are required including the laparoscopic camera. Currently these tasks are executed by a team working together. However, cooperation between two people may lower the operation efficiency due to communication errors \cite{Nurok2011} or uncoordinated motion. Three-handed trimanipulation using SRLs controlled by one operator can be a solution to the communication issue coming from multiple operators. 

SRLs can be controlled by hands-free interfaces using the human body as a reference including control from the head \cite{Kommu2007}, tongue \cite{2016tonguecontrol}, foot \cite{2019Huang}, gaze \cite{2010eyegaze} and voice \cite{YulunWangGoletaDarrinUecker2002}. Among these hands-free control strategies, the use of foot motion is an attractive source due to the foot's ability to control in multiple DoFs and its relative independence from the subject's hands and/or eyes \cite{3rdfinger, 2018Metaarm, 2019fourarm}. The analysis of subject's performance while using foot to control a SRL has only recently been considered.  

Abdi et al. \cite{Abdi2015, Abdi2016} first conducted studies of trimanual control using the foot and demonstrated that a virtual ``third hand'' can be controlled by the foot with the natural hands to head towards objects. For the given reaching task, they found that users preferred to use three hands compared to only two. The capability of subjects to use two additional arms was also considered in the later studies \cite{2018Metaarm,2019fourarm}, in which a single subject performed bipedal tel-emanipulation of two robotic arms demonstrating a quad-manual operation. These studies were either limited to the control of a single axis motion using the SRL or still in the demonstration stage without a systematic study of the user's ability for different kinds of tasks. Particularly, the studies have not yet considered tasks which required simultaneous coordination of more than two hands.

To investigate different tasks on trimanual performance, Huang et al. \cite{huangTMRB} compared the impact of the addition of a third hand to bimanual tasks involving the hands acting independently or having coupled motion. It was found that the addition of the foot controlled hand slowed operation time, however, when the hands acted independently of one another there was otherwise no reduced performance when the hands were uncoupled. The relative performance of trimanipulation was then compared to the use of multiple operators \cite{Alessia}. Three different levels of coupling between the hands were considered, where in all cases the couple outperformed a single trimanual subject. This study, however, showed evidence of an unsaturated learning effect in trimanual condition. The authors suggested that the differences in performance between the dyad and trimanipulation were the smallest for the case in which the subject limbs were continuously coupled together.

In this paper, we extend the study of \cite{Alessia} to specifically account for the effect of learning on coupled tri-manipulation. We investigated i) the learning of tri-manipulation with the addition of supernumerary limb in a three hand coordination task; ii) compared the users' performance to dyadic operation of one person performing bi-manipulation and another person using their foot, and iii) studied the effect of an unpredictable environment on the operation. A tri-manipulation teleoperation platform was built with two hand interfaces, one foot interface and a virtual reality scenario. We recruited 14 subjects for a study taking place over two days in which we asked the subjects to perform a trimanual virtual task requiring coordination of three limbs. The results show that solo trimanual performance could achieve similar level to two person's cooperation in the aspects success rate and coordination. In addition, the participants seem to have a more efficient motion in trimanual mode than when cooperating with a partner.

%figure: 
\begin{figure*}[!t]
\centering
\includegraphics[width= 0.93\textwidth]{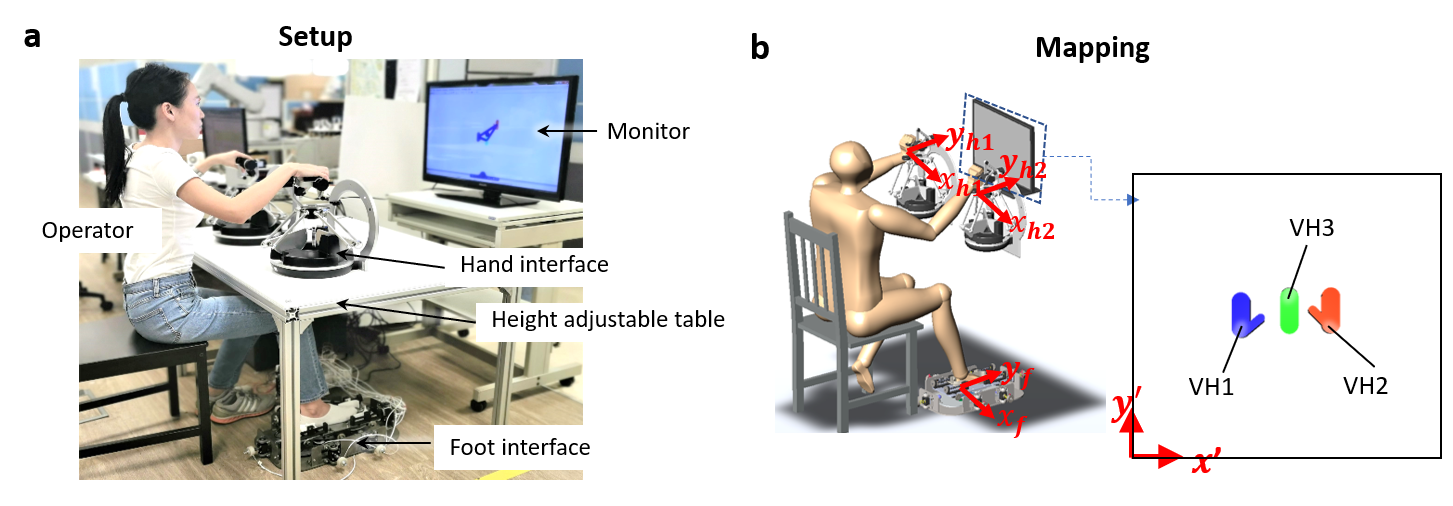}
\caption{(a) Experimental setup. (b) Mapping between hand/foot and virtual hands (VHs). VH1 (blue) is controlled by left hand, VH2 (red) is controlled by right hand and VH3 (green) is controlled by the foot.}
\label{f:mapping}
\end{figure*} 

\section{Method and material}
% human-machine-interaction
\subsection{System and task}
The trimanipulation system is comprised of two hand interfaces (Omega.7 from Force Dimension Inc., Switzerland), one foot interface \cite{2020huang}, and a virtual-reality testing environment. Fig.\,\ref{f:mapping}a shows the experimental setup with the trimanipulation system and the operator. Throughout the study, the operator sat comfortably on a chair, holding the handles of the hand interfaces, whose height could be adjusted using the mounting table, with both hands and placing their right foot on the pedal of the foot interface.  They faced the monitor which displayed feedback of the virtual task. As shown in Fig.\,\ref{f:mapping}b, the position of the hands $x_{hi}$ and $y_{hi}$ (workspace 16\,$\times$\,16\,cm$^2$) and foot \(x_{f}, y_{f}\) (workspace 4\,$\times$\,4cm\,$^2$) in the horizontal plane were used to control VHs 1, 2 and 3 respectively in the vertical $x'-y'$ plane. 
% The hands and foot positions were recorded by the interfaces were transmitted to the main computer through serial ports at a frequency of 30\,Hz. 

%figure: 
\begin{figure}[!b]
\centering
\includegraphics[width= 0.5\textwidth]{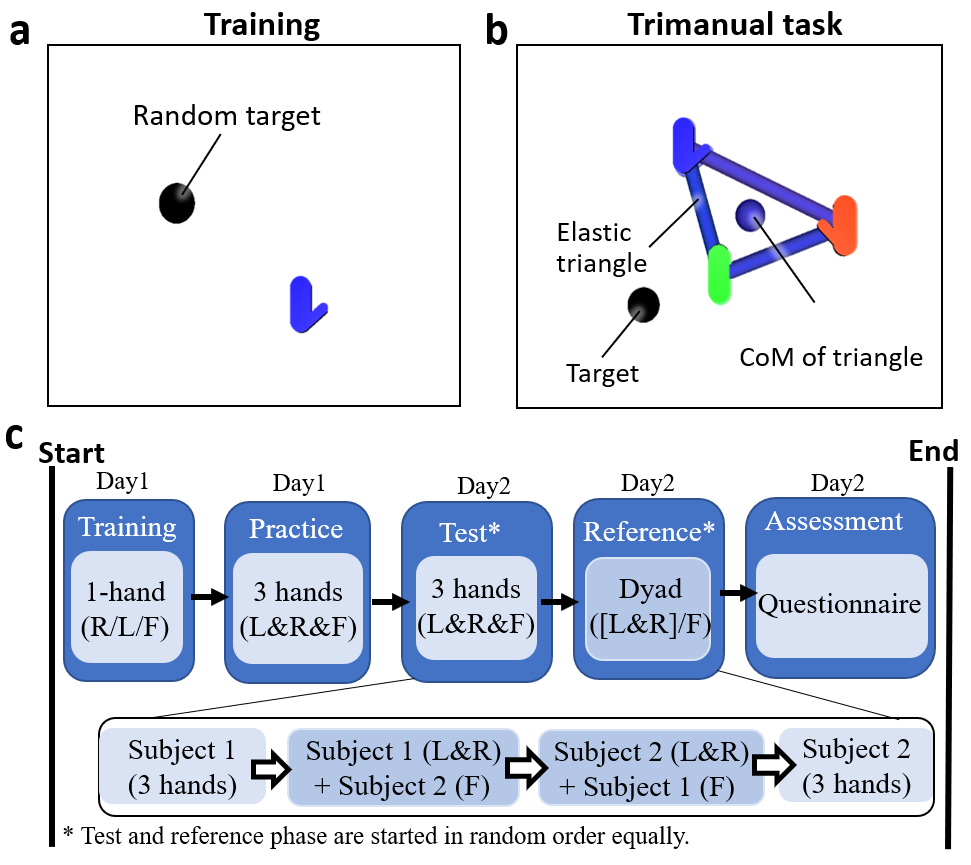}
\caption{Experimental protocol. (a) Training scene. (b) Trimanual task scene. (c) Experimental procedure flow chart.}
\label{f:protocol}
\end{figure} 
The virtual task with VHs (Fig.\,\ref{f:protocol}a) simulated a three-hand coordination scenario. This task was designed to simulate everyday activities requiring physical coordination of each hand, e.g. holding a tray or the case when assembling furniture. In the task, an elastic triangle object was held by three VHs and its center of mass (CoM) controlled by the coordinated movement of the three VHs. Haptic feedback was provided to hands to rendering the elastic force. The participants were asked to maintain the length of the three elastic edges within a set range and to move the CoM to the target. The target was updated on the screen every three seconds and changed from black to green color when it was reached. When the elastic edge was not in the required range (i.e. smaller than 0.7, or larger than 1.3 times the initial length), the elastic bar either `broke' or a `collision' was indicated through visual cues. A supplementary video is attached to show the task operation.

% %%% master-slave interaction
% \subsection{Haptic feedback}
% Both a force and visual feedback effect were imparted when manipulating the virtual object.The force of each elastic band could be calculated from position of VHs through Hooke's Law. 

%%%%%%%%%%%%%%%%%%%%%
\subsection{Experimental procedure}
Fourteen participants (eight females), without motor impairment, with an average age of $28.8\pm 2.5$ years were recruited. All subjects were right footed according to the ball-kick dominant leg test \cite{2017kickball}, and none of them regularly used any foot based gesture system. The experiment contains two days (Fig.\,\ref{f:protocol}c). 

On the first day, the subject conducted a \textit{training} phase and a \textit{practice} phase. In the training phase, the three limbs controlled a single VH one by one using either the hand/foot interfaces. A single target was shown on the screen and one of three VH were used to reach the target. This phase was designed to help the participants understand the mapping and control scheme. Each session included 20 targets. The training was stopped when the hand-controlled VHs reached 90\% of the target and the foot controlled VH reach 80\% of the targets. After the training phase, the subjects were asked to control the three VHs together and practice the trimanual task. Ten sessions of 200 trials (each session including 20 trials) were conducted in this phase. Between sessions, the subject are given 30 seconds to take a rest. 

% Different target sequences were used to verify generalisation of the results. Eight of the 14 participants used the fixed target sequence and the six subjects applied the random sequence with the same targets. In each session of random sequence, the target is shown in an unpredictable sequence which simulate a unknown dynamic changing environment.

On the second day, the subjects conducted the \textit{test phase}, which is same as the practice, and the \textit{Reference phase}, in which two subjects were randomly paired to perform the same task in ten sessions: one operator used their natural hands and their partner used their right foot to cooperatively perform the task. Half of the subjects performed the test phase first and the other half subjects completed the reference phase first. In the \textit{assessment phase}, the subject finally filled in two assessment questionnaires. We arranged the reference phase on the second day to pursue a relatively fair comparison between dyad and trimanual control modes as the tri-manipulation practice may also improve bi-manual and foot operation.

In addition, to study the sensitivity of trimanual and dyad operation to a dynamic environment (target sequence), we separated subjects into two groups with a fixed or random ordering of targets. In the fixed group, 20 targets in each session were shown in fixed order in all phases; while in random group, the targets were shown in an unpredictable sequence. Eight of the 14 participants used the fixed target sequence and the six subjects applied the random sequence with the same targets.

\subsection{Evaluation measures}
Subject trial performance was evaluated based upon different quantitative measurement types and additionally subjective measures from the questionnaire responses.
\subsubsection{Measures of performance}
\begin{itemize}
    \item \textbf{Success rate} defined as the ratio of the number of successful trials divided by the total number of trials in each session. Success occured when the CoM reached the target within a hold time that was more than 0.1s.
     \item \textbf{Efficiency ratio} was calculated by taking the real travel distance of the CoM and dividing by the straight/optimal distance to the target in successful trials. In the optimal case, the efficiency ratio is equal to 1.
    \item \textbf{Operation time} was the difference in time between the instant where the target was shown on the screen to the instant in which the CoM reach the target. The operation time is also calculated in the successful trials and normalized by dividing the optimal distance to the target.
\end{itemize}

\subsubsection{Measure of coordination}
The coordination level of every two hands was measured by the \textbf{coordination error}, which is the sum of of the dynamic deviation of each elastic band. For example, the elastic band connecting VH1 \& 2, controlled by left and right hands (L-R) is calculated through:
\begin{align}
     E = \frac{\sum_{2}^{N} {(L_{1}(n)-L_{1}(n-1))}}{\sum_{2}^{N} (P_{CoM1}(n) - P_{CoM1}(n-1)}) 
\end{align}
where $L_{1}$ is the dynamic length of the elastic band derived from the position of VH1 \& 2, $P_{CoM1}$ is the position of the CoM of the bar 1, $P_{CoM1} = (P_{VH1} + P_{VH2})/2$, $N$ is the number of recorded data point in each session, $n$ = 2,3...$N$. The coordination error of elastic bands controlled by left hand \& foot (L-F), right hand\& foot (R-F) are derived in the same way.

\subsubsection{Subjective measures}
Each participant was asked to fill a questionnaire after each control mode in the test and reference phases (Fig.\,\ref{f:questionnaire_result}). This questionnaire provided a qualitative evaluation of the efficiency, mental effort, control and coordination. At the end of the experiment, the participants were required to fill a comparative questionnaire, asking how 3-handed operation perform compared to the dyad 2-handed/foot operation using 5-Point-Likert scale in difficulty and preference. 
%figure
\begin{figure*}[!h]
\centering
\includegraphics[width=0.9\textwidth]{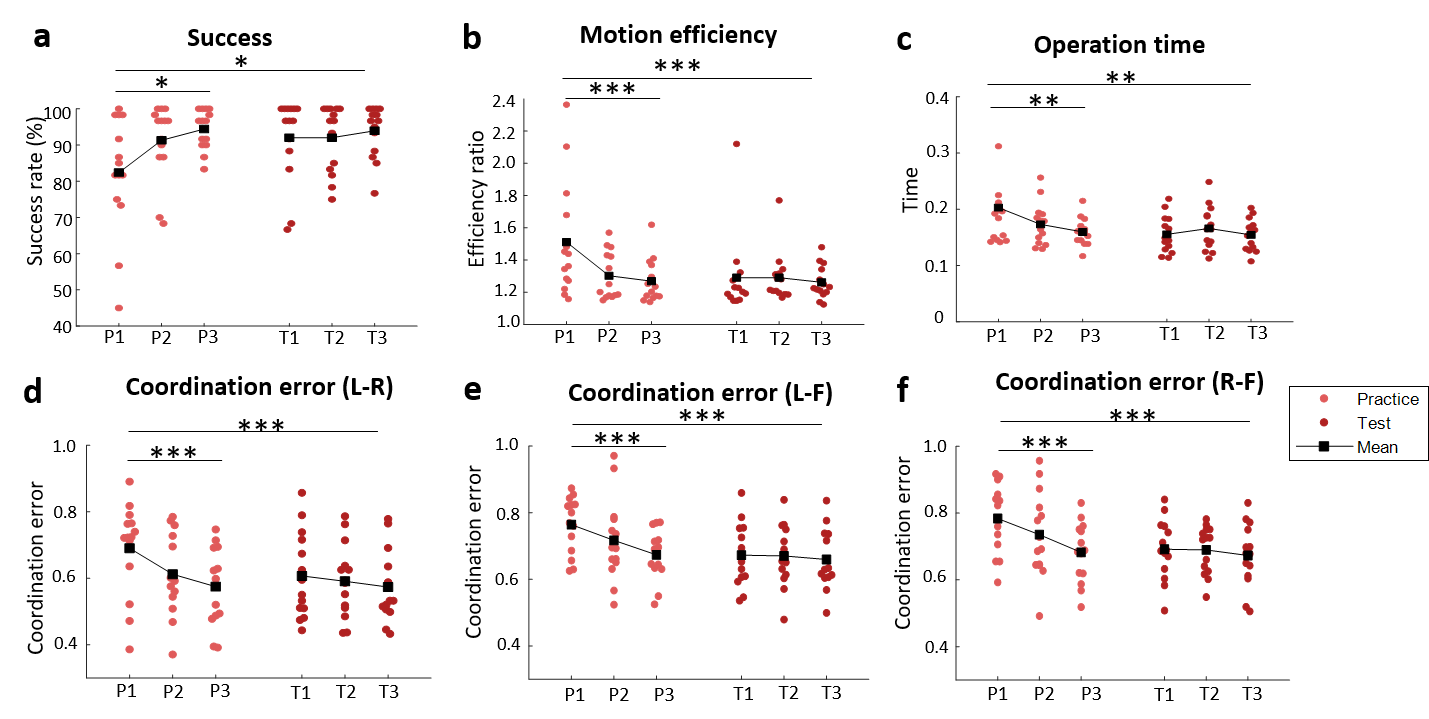}
\caption{Three-hand operation (3H) performance and coordination in practice and test phases. (a) Success rate. (b) Efficiency ratio. (c) Operation time. Coordination error of (d) left and right hands (L-R), (e) left hand and foot (L-F), (f) right hand and foot (R-F). (P1, P2, P3 are the average of measurements in sessions of 2-4, 5-7, 8-10 of the practice phase; T1, T2, T3 are the average of measurements in trials 2-4, 5-7, 8-10 of the test phase. Asterisks denote significant effects at *p$<$0.05, ** p$<$0.01, *** p$<$0.001.}
\label{f:learning_2}
\end{figure*} 

\subsection{Statistical analysis}
A Shapiro-Wilk test was conducted to examine the distribution of the data. The quantitative measurement data except success metric were found to be normally distributed. The statistical analysis was then conducted in three steps. First, the changes in the performance parameters of trimanual operation in the practice and test phases were studied. The data for motion efficiency and completion time were analyzed using one-way repeated measurements ANOVA. Custom comparisons between single conditions were conducted using post-hoc paired t-tests. The success rate (non-normally distributed) was evaluated using non-parametric one-way repeated measurements ART ANOVA with Wilcoxon signed-rank post-hoc tests. For coordination measure, two-way repeated measurements ANOVA was employed that considered changes in coordination with trials and differences between the limbs. Second, we analysed differences in success, motion efficiency and operation time between trimanual and dyad operations in the test and reference phases with two-way mixed ANOVA, the in-between factor of target sequence was also inspected. For the coordination measure a three-way mixed ANOVA was employed with an additional inter-limb predictor. All p-values of post-hoc tests in the first and second steps were adjusted with Holm-Bonferroni method to prevent type I error by multiple comparisons. Lastly, the questionnaires used an ordinal scale, therefore, a non-parametric repeated measures Friedman test was chosen for the analysis of each item. For pairwise comparisons between single conditions, the Nemenyi post-hoc test (which accounts for multiple comparisons) was employed.

% figure: 
\begin{figure*}[!h]
\includegraphics[width=0.9\textwidth]{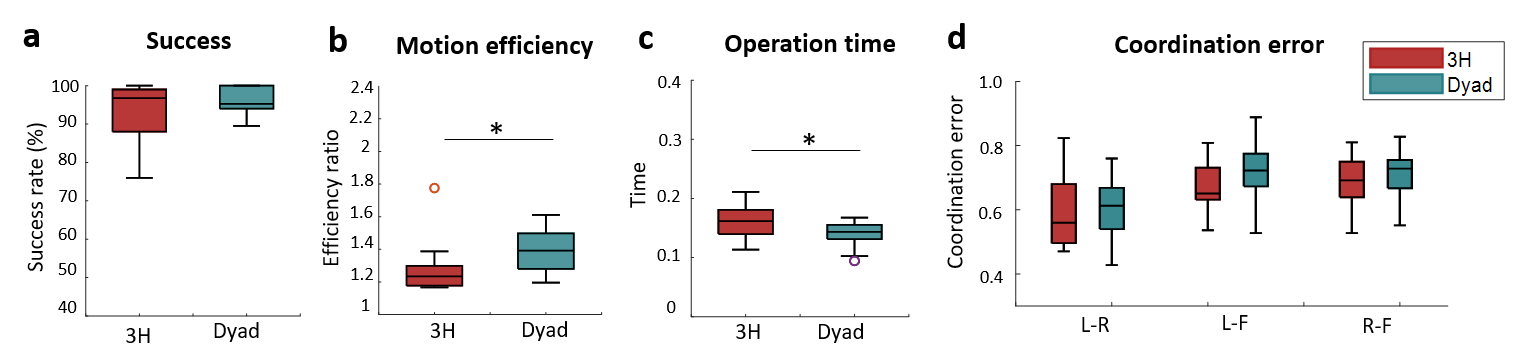}
\caption{Performance and coordination in test phase of three hands and dyad. The result is plotted in average values of ten sessions. Asterisks denote significant effects at *p$<$0.05, ** p$<$0.01, *** p$<$0.001.}
%(a) Success rate. (b) Efficiency ratio. (c) Operation time. (d) Coordination error.}
\label{f:dyad}
\end{figure*} 
%%%%%%%%%%%%%%%%%%%%
\section{Results}
% In this section, the learning effects of tri-manipulation were first investigated during practice and test phases, before the performance between the trimanual and dyadic schemes were compared in the test phase. The influence of the target sequence on the performance metrics in both trimanual and dyad operations was also analysed in addition to the result of the subject perception questionnaire. 

\subsection{Learning of the tri-manipuation}
We examined the changes in success, motion efficiency and operation time to determine whether the participants improved their trimanual performance over this time. We compared the first and last trials of the practice and test phases as well as the first and last trials of the experiment. The task success (Fig.\,\ref{f:learning_2}a) significantly improved with a growing (practice) trial number ($F(5,65) = 3.3303,\,p = 0.00971$): success at the beginning (P1) of the practice phase was significantly lower than at the end (P3) ($Z = -2.6723,\,p = 0.013$) as well as significantly lower as at the end of the test phase (T3) ($Z = -2.3231,\,p = 0.035$). However, no changes in success within the test phase were detected ($Z = -0.86809,\,p = 0.414$), so that the subjects improved regarding success mostly in the practice phase. 

We observed a similar significant tendency for motion efficiency ($F(5,65) = 6.493,\,p < 0.0001$). The participants improved mostly in the practice phase ($t(65) = 4.623,\,p < 0.0001$ for comparison between P1 and P3), but not in the test phase, where the efficiency was close to 1 during the entire phase ($t(65) = 0.541,\,p = 0.590$ for comparison between T1 and T3). Significant improvement of the efficiency between the beginning and the end of the two sessions was observed ($t(65) = 4.756,\,p < 0.0001$ for comparison between P1 and T3).

The completion time used with the trimanual condition (Fig.\,\ref{f:learning_2}c) also improved with trials ($F(5,65) = 4.255,\,p = 0.002$). Compared to the first trials (P1), the time was significantly reduced at the end of the practice phase (P3) ($t(65) = 3.460,\,p = 0.002$) as well as at the end of the test phase (T3) ($t(65) = 3.884,\,p = 0.001$). Again no improvement was observed within the test phase ($t(65) = 0.081,\,p = 0.934$ for difference between T1 and T3).

A two-way repeated measurement ANOVA yielded a significant effect of trials number ($F(5,221) = 11.963,\,p < 0.0001$), as well as limbs combination ($F(2,221) = 41.132,\,p < 0.0001$) on coordination error (see Fig.\,\ref{f:learning_2}d-f). Since interaction between both predictors was not significant, we analysed the change in coordination with respect time over all limb combinations and the differences between the limbs over all trials. Similar to the previous metrics, coordination mostly improved in the practice phase ($t(221) = 6.057,\,p < 0.0001$ for comparison between P1 and P3) and then stagnated in the test phase ($t(221) = 1.280,\,p = 0.202$ between T1 and T3). Compared to the first trials (P1) the error became significantly smaller in the last trials (T3) ($t(221) = 6.538,\,p < 0.0001$). Both conditions that involved the foot showed higher coordination error than coordination between the hands: $t(221) = 7.072,\,p < 0.0001$ for comparison between L-F and L-R, $t(221) = -8.454,\,p < 0.0001$ between L-R and R-F. No difference was found between L-F and R-F groups ($t(221) = -1.382,\,p = 0.168$). 

\subsection{Three hands vs. dyad hands/foot control}
We compared the two operation modes of trimanual and dyad control of two operators using the average result in the test and reference phases. As shown in Fig.\,\ref{f:dyad}a, no differences were found between the different conditions in terms of the task success rate  ($F(1, 12) = 1.918,\,p = 0.191$). However, the motion efficiency was closer to 1 in trimanual group, which means higher efficiency, compared to the dyad group conducting the task with a partner (Fig.\,\ref{f:dyad}b, $F(1, 12) = 6.4898,\,p = 0.026$). In contrast, the dyad control completed the task significantly faster than solo tri-manipulation controlling three limbs (Fig.\,\ref{f:dyad}c, $F(1, 12) = 5.2432,\,p = 0.041$).

Trimanual and dyad controls did not differ significantly in coordination ($F(1, 12) = 1.8643,\,p = 0.197$). The combination of different limbs resulted in significant coordination error contrasts  ($F(1.41, 16.93) = 11.788,\,p = 0.001$): movements with two hands were more coordinated than both other conditions ($t(12) = -4.987,\,p = 0.001$ for comparison between L-R and L-F, $t(12) = -3.336,\,p = 0.012$ between L-R and R-F), no significant differences between the conditions involving the foot were found ($t(12) = -0.189,\,p = 0.854$).  

\subsection{Sensitivity to target sequence}
In the same analysis with the previous section we analysed the effect of the target sequence. Using a random or fixed sequence has no impact on the success ($F(1, 12) = 0.0701,\, p = 0.796$), but significantly influenced the motion efficiency ($F(1, 12) = 13.1419,\, p = 0.003$) and has a tendency to have an impact on completion time ($F(1, 12) = 3.8793,\, p = 0.072$). The Target sequence also significantly influenced the coordination error of participants ($F(1, 12) = 22.698,\, p < 0.001$). The fixed target sequence could be achieved by subjects faster, more efficiently and coordinated. Importantly, no interaction effect between trimanual and cooperation and target sequence was found for any of metrics. 
 
\subsection{Subjective assessment}
\begin{figure}[!h]
\includegraphics[width=0.5\textwidth]{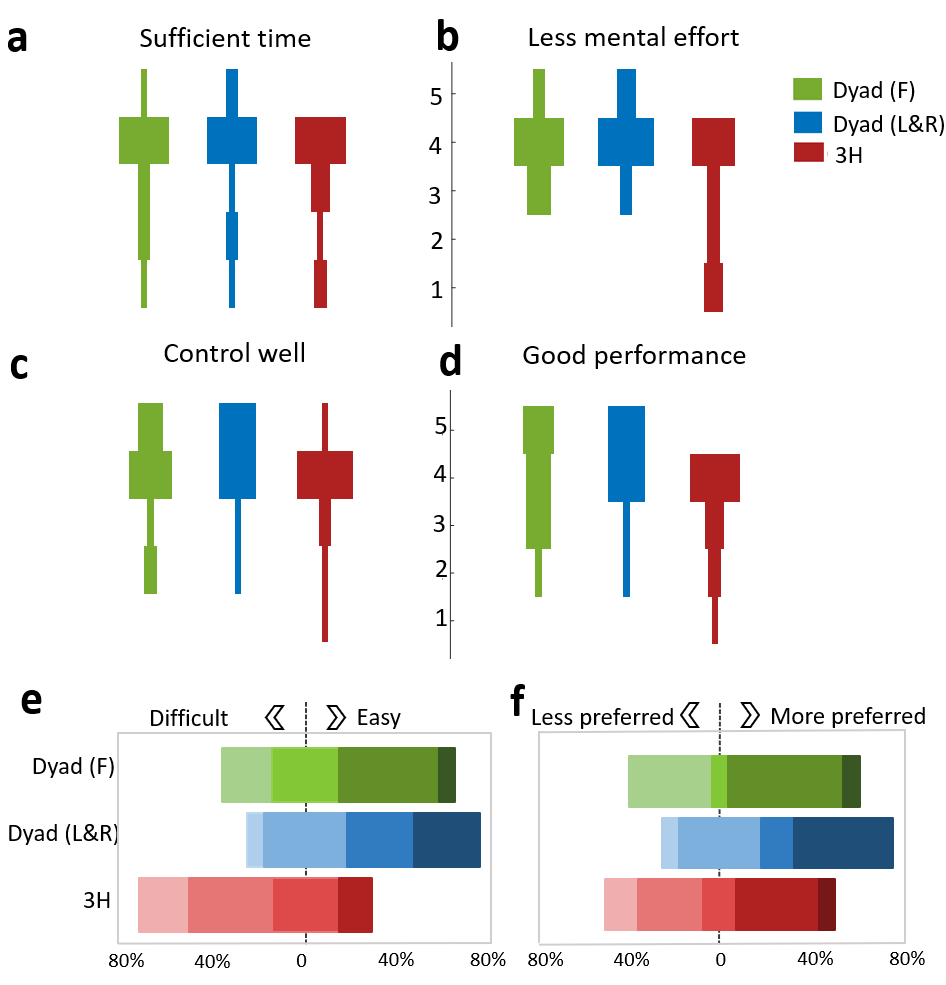}
\centering
\caption{Assessment of user experience. a-d: Histograms of assessment, where questions were answered after each section. (a) Did the subject feel they had sufficient time to complete the task, -2/+2 corresponds to insufficient/sufficient time. (b) How much mental load did the subject feel during task operation, -2/+2 corresponds to more/less mental load. (c) Agency question about whether the subject felt in control of the virtual hands (-2/+2: strongly dis/agree). (d) Did the subject feel that they were in control of the coordination of the three limbs (-2/+2: very poor/good). e,f: Final comparison of the three control modes according to (e) ``difficulty'' and (f) ``preference'', the shade of the color present the answers of from difficult/less preferred (light color) to easy/most preferred (deep color).}
\label{f:questionnaire_result}
\end{figure}

The results of the first questionnaire are shown in Figs.\,\ref{f:questionnaire_result}a-d. No significant difference was found for the time to reach the target perception ($\chi^2(2)\,=\,2.6667,\,p\,=\,0.264$). The medians for each of the three conditions were equal to 2, which indicates that the time was sufficient to achieve the task with three and two hands as well as with the foot. The conditions were perceived differently regarding required mental effort ($\chi^2(2)\,=\,13.40,\,p\,=\,0.001$). In the trimanual condition subjects reported significantly higher mental effort than when performing the task with two hands together with a partner ($p\,=\,0.029$). However, no difference regarding the perceived effort was found between the task execution with a foot and using three or two hands ($p\,=\,0.029$). 

In terms of the level of control felt for the virtual limbs, a Friedman test showed significant differences between the groups ($\chi^2(2)\,=\,6.3429,\,p\,=\,0.042$). However, Nemenyi post-hoc test did not reveal any significant pairwise contrasts between conditions (all $p\,>\,0.1$). Significant differences in perceived performance were detected ($\chi^2(2)\,=\,7.85,\,p\,=\,0.0197$): by interaction with three hands subjects rated the performance lower than when achieving the task with two hands ($p\,=\,0.048$). Other differences were not found to be significant (both $p\,>\,0.3$). 

The user assessment of difficulty and preference are shown in Figs.\,\ref{f:questionnaire_result}e,f. A Friedman test revealed significant differences for the difficulty ($\chi^2(2)\,=\,14.105,\,p\,<\,0.001$): the trimanual condition was found more difficult ($p\,=\,0.007$) and has a tendency to seem more difficult than control of the foot, but this comparison was not significant ($p\,=\,0.094$). However, a difference between the interactive conditions was not found ($p\,=\,0.612$). The conditions were also different in participants' preferences ($\chi^2(2)\,=\,11.261,\,p\,=\,0.004$). People preferred more to achieve the task with a partner, interacting with two hands than completing the task alone ($p\,=\,0.010$). While interacting with a partner, subjects tend to prefer to control two hands instead of the foot ($p\,=\,0.076$), but this comparison was not significant. 

% figure: 

%%%%%%%%%%%%%%%
\section{Discussion}
This study explored the human capability to trimanipulate in a coordination task. The experiment was conducted with a platform including two hand interfaces, one foot interface and a virtual reality scenario. The subjects were in general able to successfully perform the task, where after the trimanual group's learning that took place during the practice phases, there was no clear difference was observed between the trimanual and dyadic conditions in the test phase. This suggests that trimanipulation can be performed successful by a single subject in coupled motions.

The motion characteristics of the subjects showed a similar trend where after the initial learning similar performance was observed between the dyadic and trimanual conditions in the test phase.  There were however two points of difference: the trimanual condition produced more efficient motion, while the dyads were able to complete the task quicker.  This appears to reflect the trade-off present between the two strategies, in which the solo subject is able completely plan all of its motion allowing for more efficient overall strategies, while the dyad benefits from parallelisation of the planning which can lead to fast planning and performance.

The rankings show a clear preference for working in a dyad and the belief that trimanual operation was inherently more difficult.  This may be reflective of the dyad being more familiar and as suggested by the questionnare requiring less mental effort, or it could suggest a strong preference to the cooperation and parallelisation that the condition offers.  Despite these results, it is interesting to find that the participants did not feel that the trimanual hands-foot control requires more physical effort than the dyad. Furthermore, the subjects appears to already feel they were capable of controlling the foot and had sufficient time for its operation. These results suggest that the difference in mental is specifically associated with the need to plan for the third limb. It is unclear if further learning resulting in a more clear body representation of the foot being the third hand could help make such planning more natural or if this is a suggestion of a clear difference between the conditions.

Compared to the results of \cite{Alessia}, which considered the same coupled motion task, the findings of this study suggests that with additional sessions to account for learning to perform trimanual actions, subject can prove similarly capable to a dyad. Throughout the practice phase learning was clearly observed in all metrics. For some metrics such as the success rate and motion efficiency, this learning effect appears to have hit a clear ceiling by the test phase. However, for other metrics including the time to completion and coordination, the results show that even in the test phase there is continuing evidence that learning is taking place. It is worth noting that the metrics in which the trimanual condition gives as good as or superior performance to the dyad corresponds to those that show an apparent ceiling effect.  This suggests that future work should investigate the effect of additional sessions to account for any potential additional learning.

While the results suggest potential for trimanual performance to outperform dyads, only an abstract version of a task requiring physical coupling has been considered. Future work will therefore focus on evaluating if such results are also possible in the context of other kinds of potential supernumerary tasks or with more realistic case studies matching the presented scenario.

\section*{Acknowledgment}
We thank the subjects for joining the experiment. The experiment was approved by the Institutional Review Board of Nanyang Technological University (IRB-2018-05-051). 

\bibliographystyle{IEEEtran}
\bibliography{IEEEabrv,reference}

% Generated by IEEEtran.bst, version: 1.14 (2015/08/26)
\begin{thebibliography}{10}
\providecommand{\url}[1]{#1}
\csname url@samestyle\endcsname
\providecommand{\newblock}{\relax}
\providecommand{\bibinfo}[2]{#2}
\providecommand{\BIBentrySTDinterwordspacing}{\spaceskip=0pt\relax}
\providecommand{\BIBentryALTinterwordstretchfactor}{4}
\providecommand{\BIBentryALTinterwordspacing}{\spaceskip=\fontdimen2\font plus
\BIBentryALTinterwordstretchfactor\fontdimen3\font minus
  \fontdimen4\font\relax}
\providecommand{\BIBforeignlanguage}[2]{{%
\expandafter\ifx\csname l@#1\endcsname\relax
\typeout{** WARNING: IEEEtran.bst: No hyphenation pattern has been}%
\typeout{** loaded for the language `#1'. Using the pattern for}%
\typeout{** the default language instead.}%
\else
\language=\csname l@#1\endcsname
\fi
#2}}
\providecommand{\BIBdecl}{\relax}
\BIBdecl

\bibitem{9134968}
Z.~{Wang}, H.~{Lam}, B.~{Xiao}, Z.~{Chen}, B.~{Liang}, and T.~{Zhang},
  ``Event-triggered prescribed-time fuzzy control for space teleoperation
  systems subject to multiple constraints and uncertainties,'' \emph{IEEE
  Transactions on Fuzzy Systems}, pp. 1--1, 2020.

\bibitem{8890885}
Z.~{Wang}, B.~{Liang}, Y.~{Sun}, and T.~{Zhang}, ``Adaptive fault-tolerant
  prescribed-time control for teleoperation systems with position error
  constraints,'' \emph{IEEE Transactions on Industrial Informatics}, vol.~16,
  no.~7, pp. 4889--4899, 2020.

\bibitem{Nurok2011}
M.~Nurok, T.~M. Sundt, and A.~Frankel, ``Teamwork and communication in the
  operating room: Relationship to discrete outcomes and research challenges,''
  \emph{Anesthesiology Clinics}, vol.~29, pp. 1--11, 2011.

\bibitem{Kommu2007}
S.~S. Kommu, P.~Rimington, C.~Anderson, and A.~Ran{\'{e}}, ``Initial experience
  with the endoassist camera-holding robot in laparoscopic urological
  surgery,'' \emph{Journal of Robotic Surgery}, vol.~1, pp. 133--137, 2007.

\bibitem{2016tonguecontrol}
D.~{Johansen}, C.~{Cipriani}, D.~B. {Popović}, and L.~N. S.~A. {Struijk},
  ``Control of a robotic hand using a tongue control system—a prosthesis
  application,'' \emph{IEEE Transactions on Biomedical Engineering}, vol.~63,
  no.~7, pp. 1368--1376, 2016.

\bibitem{2019Huang}
Y.~{Huang}, E.~{Burdet}, L.~{Cao}, P.~T. {Phan}, A.~M.~H. {Tiong}, P.~{Zheng},
  and S.~J. {Phee}, ``Performance evaluation of a foot interface to operate a
  robot arm,'' \emph{IEEE Robotics and Automation Letters}, vol.~4, no.~4, pp.
  3302--3309, 2019.

\bibitem{2010eyegaze}
D.~P. {Noonan}, G.~P. {Mylonas}, J.~{Shang}, C.~J. {Payne}, A.~{Darzi}, and
  G.~{Yang}, ``Gaze contingent control for an articulated mechatronic
  laparoscope,'' in \emph{2010 3rd IEEE RAS EMBS International Conference on
  Biomedical Robotics and Biomechatronics}, 2010, pp. 759--764.

\bibitem{YulunWangGoletaDarrinUecker2002}
G.~{Yulun Wang} and S.~N. {Darrin Uecker}, ``Speech interface for an automated
  endoscopic system,'' 2002, {US Patent 6463361 B1}.

\bibitem{3rdfinger}
J.~Cunningham, A.~Hapsari, P.~Guilleminot, A.~Shafti, and A.~Faisal, ``The
  supernumerary robotic 3 rd thumb for skilled music tasks,'' 08 2018, pp.
  665--670.

\bibitem{2018Metaarm}
M.~Y. Saraiji, T.~Sasaki, K.~Kunze, K.~Minamizawa, and M.~Inami, ``Metaarms:
  Body remapping using feet-controlled artificial arms,'' in \emph{User
  Interface Software and Technology}, 2018, pp. 65--74.

\bibitem{2019fourarm}
J.~Hernández, W.~Amanhoud, A.~Haget, H.~Bleuler, A.~Billard, and M.~Bouri,
  ``Four-arm manipulation via feet interfaces,'' 09 2019.

\bibitem{Abdi2015}
E.~Abdi, E.~Burdet, M.~Bouri, and H.~Bleuler, ``{Control of a supernumerary
  robotic hand by foot: An experimental study in virtual reality},'' \emph{PLoS
  ONE}, vol.~10, p. e0134501, 2015.

\bibitem{Abdi2016}
E.~Abdi, E.~Burdet, M.~Bouri, S.~Himidan, and H.~Bleuler, ``{In a demanding
  task, three-handed manipulation is preferred to two-handed manipulation},''
  \emph{Scientific Reports}, vol.~6, p. 21758, 2016.

\bibitem{huangTMRB}
Y.~{Huang}, J.~{Eden}, L.~{Cao}, E.~{Burdet}, and S.~J. {Phee},
  ``Tri-manipulation: An evaluation of human performance in 3-handed
  teleoperation,'' \emph{IEEE Transactions on Medical Robotics and Bionics},
  vol.~2, no.~4, pp. 545--548, 2020.

\bibitem{Alessia}
A.~Noccaro, J.~Eden, G.~Di~Pino, D.~Formica, and E.~Burdet, ``Human performance
  in three-hands tasks,'' 2021. doi: 10.21203/rs.3.rs-209540/v1, preprint
  available at research square doi: 10.21203/rs.3.rs-209540/v1.

\bibitem{2020huang}
Y.~{Huang}, E.~{Burdet}, L.~{Cao}, P.~T. {Phan}, A.~H.~T. {Meng}, and
  L.~{Phee}, ``A subject-specific four-degree-of-freedom foot interface to
  control a surgical robot,'' \emph{IEEE/ASME Transactions on Mechatronics},
  vol.~25, no.~2, pp. 951--63, 2020.

\bibitem{2017kickball}
N.~{van Melick}, B.~M. {Meddeler}, T.~J. {Hoogeboom}, M.~W.~G. {Nijhuis-van der
  Sanden}, and R.~E.~H. {van Cingel}, ``{How to determine leg dominance: The
  agreement between self-reported and observed performance in healthy
  adults},'' \emph{PLoS ONE}, vol.~12, p. e0189876, 2017.

\end{thebibliography}

\end{document}